\documentclass[sn-mathphys]{sn-jnl}% Math and Physical Sciences Reference Style
\jyear{2023}%

\theoremstyle{thmstyleone}%
\theoremstyle{thmstyletwo}%

\theoremstyle{thmstylethree}%

\usepackage{comment}

\begin{document}

\title[Semantics, Ontology and Explanation]{Semantics, Ontology and Explanation}


\author[1]{\fnm{Giancarlo} \sur{Guizzardi}}
\author[2]{\fnm{Nicola} \sur{Guarino}}
\email{g.guizzardi@utwente.nl, nicola.guarino@cnr.it}

\affil[1]{\orgdiv{Semantics, Cybersecurity {\usefont{OT1}{cmtt}{m}{it}\!\&} Services (SCS)}, \orgname{University of Twente}, \city{Enschede}, \country{The Netherlands}}

\affil[2]{\orgdiv{Laboratory for Applied Ontology}, \orgname{ISTC-CNR}, \city{Trento}, \country{Italy}}


\def\EP#1{{\tt {\bf EP:} #1}}

\abstract{The terms ‘semantics’ and ‘ontology’ are increasingly appearing together with ‘explanation’, not only in the scientific literature, but also in
organizational communication. However, all of these terms are also being significantly
overloaded. In this paper, we discuss their strong relation under particular interpretations. %we present several interpretations of these
%notions with an emphasizes on the strong connection between them. 
Specifically, we discuss a notion of explanation termed \textit{ontological unpacking}, which aims at explaining
 symbolic domain descriptions (conceptual models, knowledge graphs, logical specifications)
by revealing their \textit{ontological commitment} in terms of their assumed truthmakers, i.e., the entities in one's ontology that make the propositions in those descriptions true. To illustrate this idea, we employ
an ontological theory of relations to explain (by revealing the hidden semantics of) a very simple symbolic model encoded in the standard modeling language UML%, using a methodology that we call \textit{ontological unpacking}
. We also discuss the essential role played by ontology-driven conceptual models (resulting from
this form of explanation processes) in properly supporting semantic interoperability tasks. Finally, we discuss the relation between ontological unpacking and other forms of explanation in philosophy and science, as well as in the area of Artificial Intelligence.
}




\maketitle

\section{Introduction}\label{Introduction}

The terms `semantics' and `ontology' are increasingly appearing together with  `explanation', not only in the scientific literature, but also in organizational communication. However, all of these terms are also being significantly overloaded. In what follows, we discuss a few particular interpretations of these terms with an emphasis on the strong connection between them. %To set the stage from the outset, in this paper: `semantics' refers to real-world semantics and not (primarily) formal semantics; `Ontology' (with the capital O) is  a philosophical discipline that can be employed to specify ontologies (with the lowercase o), i.e., theories/conceptualizations about what is assumed to exist in the world\footnote{we are using the term `ontology' in a philosophical sense. In Computer Science, the term was re-signified to refer to a particular computational artifact representing a conceptualization, i.e., representing an ontology in the second philosophical sense \cite{guarino1998formal}. More recently, the term has been used to refer to logical specifications in a particular computational logics. We return to this point later in the text \cite{guizzardi2007conceptualizations}.}; `explanation' concerns searching for truthmakers in one's ontology and, in doing so, revealing the real-world semantics behind a certain description (specifically, a \textit{conceptual model \cite{guarino2020philosophical,proper2021domain}}). 

In science and in engineering, models have a central role. In particular, in computer science people use symbolic models to represent their assumptions about a certain domain. These are termed \textit{conceptual models} \cite{guarino2020philosophical,proper2021domain}. A widespread requirement for conceptual models, and models in general, is that they need to have some kind of formal semantics in order to be used, and especially in order to be shared. In this paper, we will defend that, in order to be shared and integrated, beyond formal semantics, conceptual models need to be \textit{explained} in terms of their \textit{ontological commitments} to the world. This process of explanation is a process of reavealing the \textit{real-world semantics} of that model. We call this explanation process \textit{ontological unpacking}.

%In summary, the focus is on explaining a conceptual model as a (mere) description of a domain by employing tools of Ontology to reveal the ontological commitment/real-world semantics behind that description. We call this process \textit{Ontological Unpacking}%\EP{Regarding the footnote: this does make one wonder how  \emph{Ontological Unpacking} relates to \emph{Ontology Engineering} (in the non-CS sense)}.

As we shall see, building conceptual models with the goal of making explicit their ontological commitments completely changes the nature of these models. In other words, the difference between a traditional conceptual model and an ontologically unpacked version of that model is not simply one of expressivity but one of nature: while the former has a mere descriptive nature, the latter has an explanatory one. We argue that it is the latter type of conceptual models that are needed for properly supporting us in semantic interoperability tasks in critical domains (e.g., healthcare, public security and safety, finances, genomics, etc.).
%\EP{1] This resonates quite well with the line of arguing I'm testing our in (1) observing that modeling is natural, (2) this leads to modeling practices, but than (3) when things become critical we need to speak about modeling capabilities (and associated quality + RoME).}\\
%\EP{2] Note, however, that the criticality of/in a domain (health, ...) does not necessarily imply that the Semantic Interoperability tasks are critical ... you may want to strengthen that link.}

The remainder of the paper is organized as follows. In section \ref{semantics-ontology}, we discuss the relations between semantics, semantic interoperability, and ontology. Furthermore, we elaborate on the idea of ontological unpacking as explanations of conceptual models. In section \ref{analysis}, we  illustrate this idea by employing an ontological theory of relations to unpack the relations of a simple conceptual model, discussing also how the tools for ontological unpacking can be leveraged to allow for the automated visualization of the \textit{interpretations} of conceptual models. In Section \ref{more}, we briefly discuss the relations between the project advanced here and other forms of philosophical and scientific explanation, including the relations to the area of Explainable Artificial Intelligence (XAI). Finally, in Section \ref{final-considerations}, we conclude the paper by presenting some final considerations.

%\EP{Side note: In discussing semantics in the CS-sense, you could pay `political correct credits' by using (a modified version of) the statement that I've been using: (1) Logic has brought discipline to the way we reason (and accidentally has enabled us to reason about the `semantics' of programs), while (2) Conceptual Modeling (and/or Semantic Interoperability) brings discipline to identifying what we reason about.}

%massively overloaded in Computer Science over the years. 
%- Right to Explanation
%- Models of Different Nature

\section{Semantics, Ontology, Explanation}\label{explanation}
\label{semantics-ontology}
%\EP{LaTeX alert: using the explanation label twice. Both at section and subsection level}

\subsection{Real-World Semantics and ontologies}\label{explanation}

Semantics is a function. More specifically, it is a function mapping elements in a domain of symbols to a \textit{semantic domain}. Syntax, and by that we mean abstract syntax, defines a structure that delimits all the expressions that are well-formed according to a given grammar. In computer science, they are typically expressed as BNF specifications \cite{mccracken2003backus}, graph grammars \cite{zambon2013abstract,zambon2017formal}, abstract syntax trees \cite{fischer2007abstract}, or metamodels \cite{bezivin2006model}. But what exactly is a semantic domain? 

When computer scientists talk about semantics they are typically referring to either operational semantics or denotational semantics. Operational semantics translates %(\EP{i.e. the translation is the function as mentioned above})
a domain or problem specification to another specification that can be executed by a computer, so that its semantic domain is an executable specification; %(\EP{Think you need an `or' here ... the word `either' calls for an `or' down the road ;-)})
denotational semantics maps %(\EP{i.e. the semantic function}) 
a symbolic structure (a syntactic artifact) to another symbolic mathematical artifact, so that its semantic domain is a symbolic structure that can serve to provide an interpretation for the original symbolic structure. An example of denotational semantics is the so-called Tarskian Semantics/Model-Theoretical Semantics \cite{tarski1983logic,feferman2006tarski} that one finds in (computational) logics. In this case, a semantic domain is a set-theoretical structure (containing sets of individuals), and an interpretation is a function mapping non-logical constants appearing in logical specifications to those sets. Semantics, in this sense, is a way to provide a univocal interpretation (again, in a set-theoretical sense) to symbols in a logical language. Nothing more, nothing less. 

This is all fundamental for mathematical and computational logics. However, it is not the type of semantics we discuss here. Our focus is on something that is prior to that, and that should always inform these set-theoretical structures and their constraints. This is the sense of semantics that is meant by philosophers, linguists, cognitive scientists and, frankly, everyone else outside logics and computer science %\EP{See our earlier point regarding your aim to `learn CS people a lesson'. Depending on that stance, you may want to use more `observational' words, and simply observe that CS has taken a rather limited view on semantics}. 
%\EP{Also, ... the CS-semantics ... brings discipline to how we `compute' ... what is missing is discipline on defining what we `compute about' ...; This could be a more `constructive' way to signal the need for a complementary view on CS-semantics.}

\begin{figure}
    \centering
    \includegraphics[width=.8\textwidth]{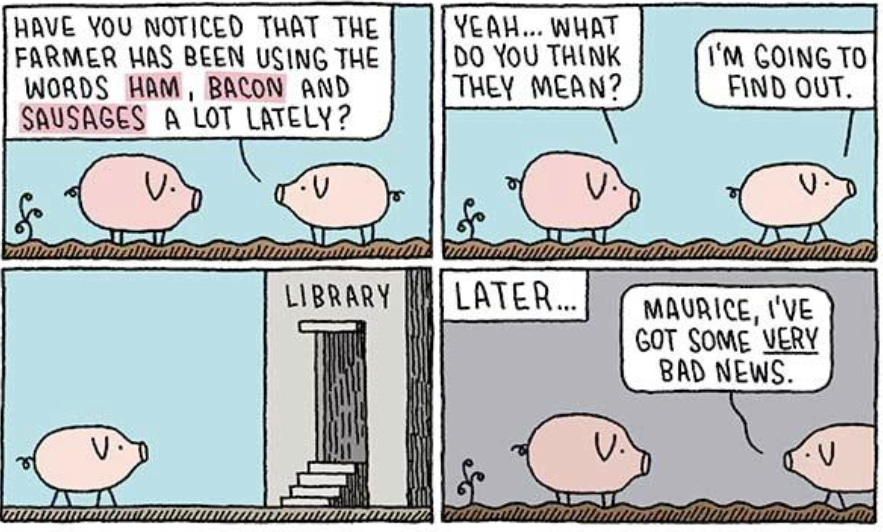}
    \caption{A Cartoon about Pigs (By Tom Gauld \copyright)}
    \label{fig:pigs}
\end{figure}

To illustrate this sense of semantics, let me employ the cartoon by the great cartoonist Tom Gauld, which is reproduced in Fig.\ \ref{fig:pigs}. Mapping the words Ham, Bacon and Sausages to subsets of a semantic domain would not do Maurice here any good, 
and would not \textit{explain} why having these words frequently mentioned by the farmer would be very bad news for the pigs. Obviously, and for the same reason, it would not explain what kind of reasoning steps we take to understand the implicit joke. This cartoon has everything we need to start making the connection between semantics, ontology and (a form of) explanation. 

In order to understand the story in this cartoon, one needs to understand (at least) that: (1) ham, bacon and sausage are man-made food products that are constituted by quantities of pork, i.e., pig's meat; (2) quantities (in the technical sense of \cite{guizzardi2010representation}, i.e., amounts of matter) of pork bear a \textit{historical dependence} relation to individuals of the type Pig \cite{fonseca2019relations}; (3) Maurice and his swine friends are instances of the type Pig. In other words, one needs to understand a system of entities (types, individuals) and relations (e.g., of parthood, historical dependence, instantiation) \cite{fonseca2019relations,guizzardi2005ontological} that exist out there in the world, i.e., beyond the descriptions and utterances we can make about this part of reality\footnote{By world and reality, we mean conceptualizations thereof, i.e., world/reality as conceptualized by cognitive agents \cite{guizzardi2007conceptualizations,guizzardi2022ufo}.}. %{\color{red}RF: or in what one believes the world to be (or you want to take necessarily a realistic stance?)}
This system of categories and their ties in the world, of which these descriptions/utterances are about is what an \textit{ontology} is. 

This ontology (of pork, pork products, pigs, etc.) reflects a semantic domain that is somehow anchored to the `real' world, or at least to the ordinary world as filtered and organized by human perception and cognition. The type of semantics (again, as a function) that maps symbols to the entities (types, relations, objects, events, aspects) that are part of this ontology is what we will call (for the sake of historical compatibility) \textit{real-world semantics} \cite{wieringa2011real,guizzardi2020ontology} but we could equally call it \textit{ontological semantics}.

Note that here we are talking about ontology with a lower-case `o'. As such, an ontology can be defined as a specific theory about the kinds of entities and their ties that are assumed to exist by a given description of reality \cite{guizzardi2020ontology,guizzardi2007conceptualizations}. This is a sense of ontologies that appears in the first mention of the term in Computer Science, namely, in the visionary paper entitled ``Another look at data" \cite{mealy1967another} by George Mealy. Mealy then  writes: \textit{``data are fragments of a \textbf{theory of the real world}, and data processing juggles \textbf{representations} of these fragments of theory... The issue is one of ontology or the question of what exists"} (our emphasis). In this passage, Mealy makes a reference to ``On what there is" \cite{quine1948there}, in which the philosopher W.V.O. Quine -- whose definition of ontology is akin to the one we just mentioned -- elaborates on the notion of \textit{ontological commitment} of a description, i.e., what entities or kinds of entity that description must be capable of referring to if its affirmations are true.

\begin{figure}
    \centering
    \includegraphics[width=.8\textwidth]{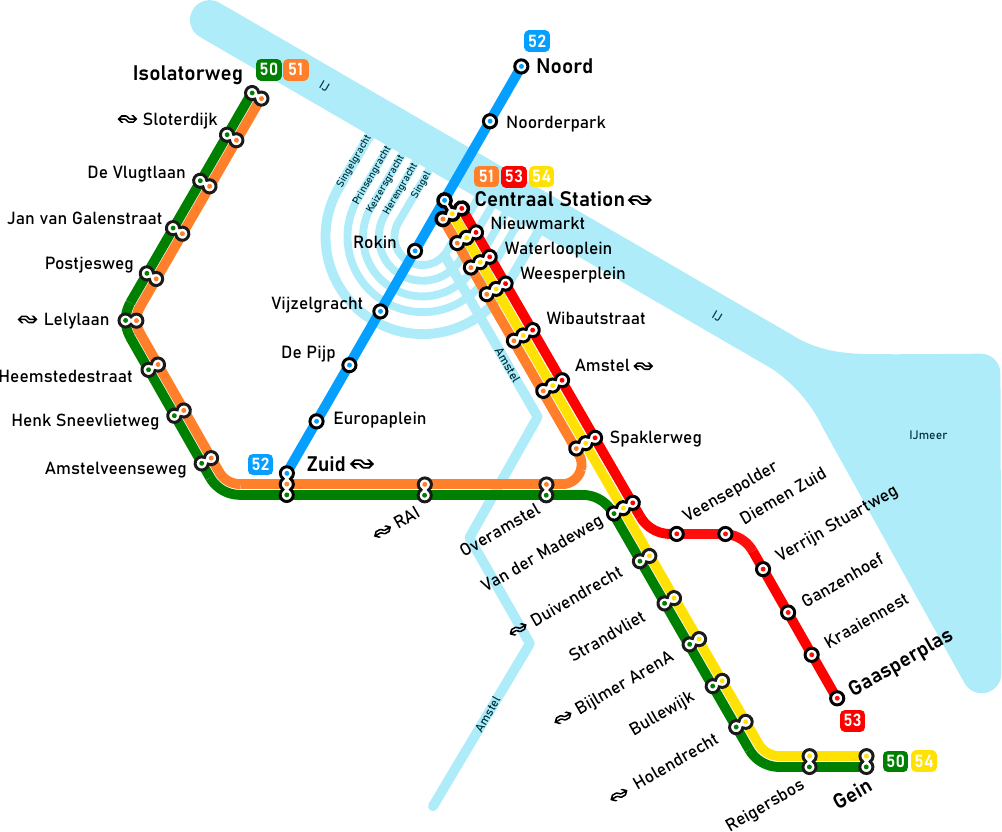}
    \caption{The subway map of Amsterdam (by Michiel Straathof).}
    \label{fig:subway}
\end{figure}

Now, any symbolic representation that has real-world semantics, i.e.\ that is not just a piece of mathematics, makes an ontological commitment \cite{guizzardi2020ontology}, i.e., commits to a given worldview, a given ontology. Let me make another example outside computer science. Consider the subway map depicted in Fig.\ref{fig:subway} -- the subway map of the city of Amsterdam. This map commits to a worldview in which there are subway lines that: have directions, intersect each other, are composed of subway stations, etc. To understand this description is to understand these entities and the relationships among them.  %\EP{The map does not make the ontological commitment that the subway lines have a specific geographic positioning, only a relative positioning is assumed ...}

Moreover, this map can be decomposed into a set of propositions (e.g., De Pijp station is part of line 52 and lies between the Europaplein and Vijzelgracht stations, lines 50, 51 and 52 intersect in the Zuid station, etc.). What makes these propositions true (if they are indeed true) are entities in the world bearing certain properties and certain relations to each other. The former are called \textit{truth-bearers} (i.e., entities to which truth-values can be assigned); the latter are called \textit{truthmakers} \cite{moltmann2007events,guarino2018reification,guarino2019weak}. These truthmakers hold explanatory powers and, in fact, revealing the ontology behind a description is a fundamental type of explanation. 
%\EP{This does mean that our ontological commitment when regarding the metro network, it would need to cover both the network and its truth-makers ... which is probably explained below ;-) }

In section \ref{explanation}, we discuss this particular notion of explanation but, for now, it suffices to appreciate the following example (still referring to fig. \ref{fig:subway}). Suppose one wants to go by subway from Sloterdijk to De Pijp. An explanation for \textit{why}\footnote{See the notion of why-questions in \cite{van1977pragmatics}.} one could ``take line 50 direction Gein, and switch to line 52 direction north (Noord) in Amsterdam Zuid station" derives its explanatory powers from the denizens of our ontology, i.e., those lines, stations, and their relations. As put by the philosopher T. Y. Cao \cite{cao2003ontological} when referring to scientific descriptions and their explanations: \textit{``the notion of a basic ontology in a scientific theory refers to the irreducible conceptual element in the logical construction of what is assumed to really exist in the domain under investigation. As a representation of deep reality, the basic ontology enjoys a great explanatory power. That is, all appearances should be derivable from it as a result of its behavior."} %\EP{Would expect that for ``explaining'' there is also a need for a stop criterion ... how much detail is needed?
%For each truthmaker we can wonder why it is true, requiring yet other truthmakers in the unpacking. So, where to stop?}

\subsection{Semantic Interoperability and Ontology}\label{explanation}
Why should computer scientists be interested in semantics in the aforementioned sense? The first direct answer is that this is the only way the symbols manipulated by information systems acquire meaning. In other words, computers are symbol-manipulating machines but the ascription of meaning to these symbols so that they can represent entities in the world (e.g., real objects, events, relationships) is not done by the computer itself. It is us who have to translate our conceptualizations of reality to symbolic representations, which are then manipulated by machines producing new symbolic representations (as well as interventions in the world%\EP{would make this point more elaborately ... drones and robots. And would then add the observation that in reporting their actions, they do so in symbols again}
). It is also us who have to interpret these resulting symbolic structures %(\EP{and the interventions we observe such machine to perform in the world around us}) 
in terms of entities in our semantic domain. So, doing ontology work is absolutely inevitable for any discipline dealing with information. 
%\EP{Thinking a bit more about the last remark made above. Would suggest to state that in the case of ``computing machinery'' (also the CM of ACM ;-) ) there are at least four ``spheres'' where information is created/used/interpreted:
  %(1) requirements engineering; i.e. discussing the required properties in relation to the domain it needs to operate in/for, 
  %(2) system design; the design of the system,
 % (3) system operations; i.e. the operational system and how it exchanges information with its users and environment, and 
 % (4) explaining/motivating the actions as undertaken by the system. 
  %``Sphere'' 3 becomes particularly important in the case of AI, complex interconnected systems, robotics, etc, also in relation to ethical considerations of course. 
 % For each of these spheres, the ontological commitments needs to be made explicit.
%} 

However, from a practical point of view, this need is even more acute nowadays because almost all of the questions we need to have answered in government, science, organizations, can only be answered if we put together data that now reside in autonomously developed data silos \cite{waters2009global,torab2023interoperability,guizzardi2020ontology,jaulent2018semantic,mccomb2019data}. For example, suppose we want to systematically answer the following question, by executing only one query ($Q_1$): \textit{which organizations have a contract with a governmental institution and have donated money to the political campaign of any politician that governs that institution or who has a personal relationship with someone that does?} 

In order to answer this question, there is obvious semantic investigation work that we need to do. For example, we need to understand that, if an organizational unit B is part of an organizational unit A, then whoever governs A then also governs B (due to the semantics of hierarchical relations). Moreover, since parthood is a transitive relation, then any part of B is also part of A transitively and, hence, all these parts are ultimately governed by whoever governs A. Additionally, one has to understand the semantics of \textit{having a personal relationship}. Finally, one has to understand a plethora of temporal aspects including being able to answer in which period: (a) campaign donations are made; (b) contracts are established between (potentially donor) companies and organizational units; (b) a certain organizational unit is part of another; (d) people are assigned to play certain roles in organizational units; (e) personal relationships (whatever those mean) hold. Understanding (a-e) requires \textit{explaining} what makes the relations in this scenario hold between those specific relata when they do, i.e., what are the truthmakers of these relational propositions. 

Besides answering questions (a-e) in separation, perhaps, the most challenging aspect of answering $Q_1$ is the fact that the data needed for that typically exist scattered in several different information systems that use different vocabularies and data structures (e.g., one dealing with organizational structures, one dealing with people allocation, one dealing with campaign donations; one dealing with governmental contracts, several different systems dealing with personal relationships, for example, notary systems, chamber of commerce systems, social networks, etc.). Now, connecting these different systems/silos requires finding out how the references in the world represented in these different systems/silos relate to each other. In other words, it requires us to precisely understand the different worldviews embedded in these systems, or the different ontologies they commit to. This is exactly what \textit{semantic interoperability} refers to \cite{guizzardi2020ontology}, and is one of the biggest problems that must be solved in the next decade \cite{mccomb2019data,guizzardi2020ontology}. 

For example: if one has \textit{person}, \textit{contract}, or \textit{economic transaction}, represented in systems $S_1$ and $S_2$, can one assume that their meaning is the same for all these different systems? If so, what does that entail? If not, what are the alternative relations that can hold between their referents? 

In case, e.g., the meaning of \textit{economic transaction} in $S_1$ ($ET_1$) is identical to that of the homonymous term in $S_2$ ($ET_2$), then we have that whatever property holds for $ET_1$ holds for $ET_2$, but that property holds for whatever other notion of economic transaction that is identical to $ET_2$ in any other system. This is because identity is a very strong relation -- it is an equivalence relation (i.e., reflexive, symmetric and transitive relation) and also a relation that obeys the so-called Leibniz Law \cite{feldman1970leibniz}, i.e., if two things are identical then they necessarily have the same properties (in a modal sense). 

Now, if the meaning of $ET_1$ is not identical to that of $ET_2$, then what are the possible relations between them? Maybe specialization? Sibling subtypes of an implicit common supertype? Dependence? If the latter, what kind of dependent? Existential? Generic? Historical? Notional? Future? In order to answer all these questions, we need a conceptual toolbox that supports a systematic process called \textit{ontological analysis} \cite{guarino2005applied}. Ontological Analysis can help us to: (i) identify and analyze those truthmakers, thus understanding their nature; (ii) ask and answer systematic questions that can help to identify how the notions in these different ontologies relate to each other, and which kind of ontological relations exist between them; (iii) understand the formal meta-properties entailed by these ontological relations. To perform (i-iii) in a systematic way, we need the support of \textit{Ontology} (now with the capital `O'), which is a branch of philosophy devoted to developing general theories about the nature and structure of the world, as well as methods for performing proper ontological analysis. 

\subsection{Ontology and Explanation}\label{explanation}

The project we outline here shares some tenets with the notion of \textit{Ontological Explanation} defended by T. Y. Cao, who advocates that \textit{``whenever we have something important but difficult to understand, we should focus our attention on finding what the primary entities are in the domain under investigation ... Discovering these entities and their intrinsic and structural properties, rather than manipulating uninterpreted or ill-interpreted mathematical symbols, or speculating on free-floating universal laws and principles, is the real work of science. Mathematical formalisms and universal laws and principles are relevant and important only when they have a firm ontological basis."}.

The project is, however, also much more humble in two important ways. Firstly, the primary role of Ontology here is \textit{descriptive} (as opposed to \textit{revisionary}) \cite{guizzardi2007conceptualizations}. Thus, its primary goal is not so much to reveal the ultimate structure of the world but to reveal and make explicit the content of human communications. In this enterprise, \textit{“[t]ables, detente, machinists, and love affairs are absolutely not merely epistemic entities ... all these entities are things beyond us, things in the world. [They] are not ‘ways of taking the world’, They are the world taken a certain way.”} \cite{smith2019promise}. In other words, the project is to systematically reveal the ontological commitment underlying a given symbolic description, or to reveal that ‘world taken a certain way' (i.e., from a given perspective, granularity, etc.). %\EP{See our earlier remark about the potential to have a `stop criterium'} 
The entities composing that revealed ontology are the truthmakers of the propositions composing that description. So, in this sense, this notion of explanation is much more strongly related to the notions of grounding \cite{thompson2016grounding} and truthmaking \cite{smith2007truthmaker} in the literature of \textit{metaphysical explanation} \cite{brenner2021metaphysical}. Second, since our main objective is to contribute to the area of conceptual modeling in computer science, our focus is on a particular type of symbolic artifact, namely, conceptual models \cite{guarino2020philosophical}.  

Conceptual models have traditionally been employed to capture and represent the main concepts that exist in a domain, using a level of abstraction that is suitable to develop an information system following users’ requirements. In this classical sense, a conceptual model provides an \textit{information structuring function} for a given application domain%(\EP{Well ... in the paper by Nicola, You and John, you do trace the history of CM beyond IS design ...})
. As such these models are merely descriptive%(\EP{Not sure we understand this, as they are mostly used for revisionary purposes ... i.e. IS development. Even more, often this also leads to a revision of the UoD, as part of a general redesign ...})
, i.e., they are focused on representing truth-bearing propositions. Now, in order to fully support semantic interoperability %(\EP{maybe add a statement, linking it back to the motivations as discussed above ...}) 
in complex critical domains, we need conceptual models that provide conceptual clarification and strive to minimize ambiguity in communications regarding the nature of entities and their connections, which are assumed to exist in a given domain. We call this the \textit{ontological function} of a conceptual model. In this sense, the model must strive to represent the exact intended conceptualization (that is, the exact set of possible interpretations) of the domain that it is intended to represent. In other words, the model should be as much as possible explicit and transparent with respect to its ontological commitment/real-world semantics. 

Again, revealing these semantics of an information artifact is a fundamental type of explanation for symbolic models. In Latin languages such as Portuguese, Italian, Spanish, and French, the terms for explanation literally mean “to unfold” (or to unpack). Thus, we use the term \textit{Ontological Unpacking} to refer to a process of ontological analysis that reveals the ontological conceptual model (a conceptual model in its latter function) behind an information structuring conceptual model (the conceptual model in its former function). 

In a series of papers, one of us has conducted with colleagues the ontological unpacking of a virus conceptual model \cite{bernasconi2022semantic}, and of a model representing a part of the human genome dealing with metabolic pathways \cite{garcia2022ontological}. In the next section, we will illustrate the approach by briefly presenting an \textit{ontological theory of relations} \cite{guarino2015we,fonseca2019relations,guizzardi2005ontological}, and by showing its power in revealing the semantics of several relations.     
%\EP{It remains a bit unclear (in the text) how exlanation(s) in an AI sense related to the explanations needed for CMs. }

%\EP{Also wondering if we should make a difference between (1) explanation of causalities, (2) explanation of decisions (sometimes based on causality), and (3) deconstruction of concepts and relations (typically driven by a desire for ontological in(tro)spection; curiosity). Each of these will require (differing levels of) ontological unpacking, with associated stopping criteria.}

\section{Ontological Unpacking}
\label{analysis}

\begin{figure}
    \centering
    \includegraphics[width=.7\textwidth]{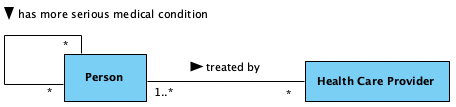}
    \caption{A simple UML model describing a relation between Healthcare Providers and people treated by them.}
    \label{fig:simpleuml}
\end{figure}

In order to illustrate the notion of ontological unpacking we have been discussing in this paper, let me start with a very simple UML conceptual model (Fig.\ref{fig:simpleuml}) but one that is representative of the nature of models one finds in practice. This model might serve well the purpose of a guiding blueprint for information structuring. However, in what sense can it explain what is going on here? Unless one has prior 
knowledge of the semantics of these types and relations, this semantics is not made explicit by the model itself. For example: what does it mean for a person to be treated by possibly several healthcare providers? In the same treatment? In several treatments? Synchronically? Diachronically? Mutatis mutandis, what does it mean for a Healthcare Provider to treat many people? Again, in the same treatment? Different treatments? Through time? Moreover, how can we explain that those persons who are in a more serious condition than others  require precedence in their treatment? 

\subsection{Analyzing the relations}
We can answer these questions in a systematic manner. Let us start by looking at the relations that appear in the model. If a person \textit{has a more serious medical condition than} another person, what would make that proposition true? In other words, what is the truthmaker of that relational proposition? Suppose that people can have certain medical conditions (diseases and disorders) and that these health conditions can be classified in a scale of severity. So, \textit{person x has a more serious medical condition than person y iff both x and y have medical conditions and x has at least one medical condition that is more severe than all the medical conditions of y}. Even this toy example with a radically simplified assumption already tells us many things. 

Firstly, the relation of \textit{has a more serious medical condition than} that allegedly holds between x and y is not a `genuine' %\EP{(truthmaker)} 
relation between them at all. 
%\EP{(Here you would need to clarify what qualifies as a relation. With an ORM perspective, we would argue that the latter relation between $x$ and $y$ does exist, but that it might be derived. Which is fine. It is just not `basic' ... and indeed ... does not immediately reflect the (underlying) truthmakers. Would make this a bit more explicit here, especially since it is an illustrating example ;-)}
Instead, this relation is a derived one from another relation of \textit{is more severe than}  holding between two medical conditions of x and y. In other words, the former relation is actually completely reducible to the latter, so that medical conditions and not people are the interesting relata here. 

%\EP{A question from a database design point of view ... even though an `unpacked' conceptual model may involve a relation that us unpackable, one might decide for system design considerations not to unpack this relation, but rather accept that e.g. the truth is accepted by `reporting'. For instance, when a doctor states that $x$ \textit{has a more serious health condition than} $y$, one may (for practical purposes) accept this as an immediate truth towards the information system. Of course ... if the doctor is asked to \emph{explain} this statement, we end up drilling things down to the underlying actual truth makers (with some stopping criterion ...).}

Moreover, medical conditions are intrinsic dispositions of people in a given point in time, so the alleged relation of \textit{has a more serious medical condition than} has no relational content in the ordinary sense. Stated differently: if Paul \textit{has a more serious medical condition than} John, we can infer no real connection between Paul and John. In fact, if Mary in the other side of the world exhibits now a health condition that is more severe than all the medical conditions of Paul then this relation is immediately established (just by the sheer existence of those medical conditions and their severity values) between Mary and Paul but also between Mary and John, due to the transitivity of that relation. In fact, why is this relation transitive to begin with? 

%%%% EP here I'm at on Monday at 06:40

Well, the relation of \textit{is more severe than} is again derived from the relation of ordering between severity levels in that \textit{conceptual space} \cite{gardenfors2004conceptual}%\EP{(Here we wonder if our earlier remark regarding the database and the decision to record the derived relation, actually implies a different conceptual space. Thus ... also suggesting a hierarchy of conceptual spaces ... some of which are more "implementation oriented" (but not in the utility design sense). This also suggests a potential relation to a paper by Terry and me regarding ``database optimisation before sliding down the waterfall'' of database design)}
. Since (a) \textit{has a more serious medical condition than} is derived from (b) \textit{is more severe than}, which is in turn derived from (c) the ordering relation between severity levels, the meta-properties of relations (a) and (b) are also derived from the ultimate grounding relation (c). In other words, (a) and (b) are totally ordered relations (i.e., non-reflexive, asymmetric, transitive and total) because the underlying severity space here is a totally ordered space (isomorphic to the natural numbers ranging from 0 to 100). Furthermore, one should notice that the relations like (c), within the elements of a conceptual space, belong to completely different categories of relations: their truthmakers of (c) are the relata themselves, since the relation holds in virtue of their intrinsic nature. Thus, it is the sheer existence of the relata that makes the relation hold between them. 

Relations like (c) are called \textit{Internal Relations} in \cite{fonseca2019relations}. Relations like (b) and (a), in contrast, are  called \textit{Comparative Relations}. In these cases, their truthmakers are the qualities of intrinsic aspects of their relata (the severity qualities of medical conditions in case (b); the medical conditions themselves in case (a)). 

On another note, health conditions are not \textit{essential} aspects of people, i.e., people can exist without bearing medical conditions. So, we can differentiate between the notions of \textit{Healthy Person} and \textit{Unhealthy Person} (a person bearing at least one medical condition), and relation (a) should actually hold between the latter. These two types classify individuals of the same kind (Person) but they are not themselves kinds. They are dynamic types defined in terms of intrinsic dynamic classification conditions. These are called \textit{phases} \cite{guizzardi2005ontological}. To put it simply, instances of the kind Person can move in and out of the extension of these phases while maintaining their identity, and they do that due to changes in their intrinsic aspects. For example, people move in an out of the phase Unhealthy Person because of changes in their medical conditions. 

Now, let us move on to the relation \textit{being treated in}. This relation belongs to yet another ontological category, namely, that of \textit{Material Relations} \cite{fonseca2019relations,guarino2015we,guizzardi2008s}. These relationships (i.e., instances of Material Relations), unlike (c), cannot be explained by the sheer existence of the relata, e.g., if John is treated by the Twente General Hospital (TGH) it is not simply because these entities exist. Additionally, unlike (b) and (c), if this relationship holds it is not reducible to \textit{intrinsic} aspects of John and TGH. Instead, for this relation to hold, something else in the world must exist, something truly relational binding these relata. This entity is called a \textit{relator} \cite{guizzardi2005ontological,fonseca2019relations,guizzardi2008s}. A relator is a bundle of relational aspects (qualities and dispositions \cite{guizzardi2022ufo}) that: (i) inhere in a relatum; (ii) are \textit{specifically dependent} on the other relatum\footnote{For simplicity, we are restricting the discussion here to binary relationships.}; (iii) are \textit{historically dependent} on the same foundational event \cite{guizzardi2005ontological}. In legal contexts, these relational aspects constituting relators typically include commitments, claims, liabilities, etc., that entities bear w.r.t. each other in that scope \cite{griffo2021service}. A relator existentially depends on a multitude of entities, thus, binding them. The material relation then holds in virtue of the existence of that relator. Given that relators are bundles of relational aspects, material relations have true relational content. In this example, the relator at hand is a \textit{Treatment}. 

\subsection{Clarifying cardinality constraints}
\begin{figure}
    \centering
    \includegraphics[width=\textwidth]{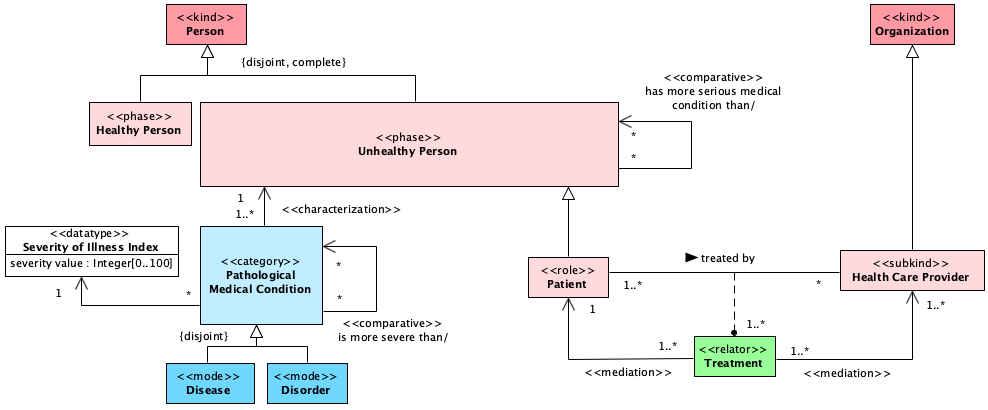}
    \caption{The result of an ontological Unpacking of the model of Fig.\ref{fig:simpleuml}.}
    \label{fig:ontouml-model1}
\end{figure}

In Fig.\ref{fig:ontouml-model1}, one can observe how the explicit representation of the Treatment relator disambiguates the (necessarily) underspecified cardinality constraints of the original relation\footnote{This model is represented in the OntoUML language \cite{guizzardi2005ontological,guizzardi2021types,almeida2019events}. OntoUML is a redesigned version of UML that has its modeling primitives and semantically-motivated syntactical constraints derived from the ontological distinctions and axiomatization put forth by the foundational ontology UFO \cite{guizzardi2022ufo}.}. The original constraints simply stated that a Healthcare Provider could treat many people, and that people might be treated by several Healthcare Providers. However, this would be compatible with: people being able to participate in many treatments, each of which involving a single Healthcare provider; people participating in one single treatment possibly involving multiple Healthcare providers; people participating in many single treatments possibly involving multiple Healthcare providers; Healthcare Providers participating in one single treatment possibly involving multiple people; Healthcare Providers participating in multiple treatments possibly each involving a single Person; Healthcare Providers participating in multiple treatments each possibly involving multiple people. This is the so-called problem of \textit{collapsing single-tuple and multiple tuple cardinality constraints} \cite{guizzardi2005ontological}. Without the explicit representation of relators, the ambiguity connected to this problem is present in all material relations\footnote{Note that the problem of cardinality constraints mentioned above concerns only material relations. For example, if we state that a medical condition is more severe than many other medical conditions, there is no cardinality ambiguity there -- it simply means that the former has a severity value that is higher than those of these other conditions.}. On the contrary, Fig.\ref{fig:ontouml-model1} shows how this ambiguity can be easily controlled by changing the cardinality constrains of the mediation relation: in this case, a Treatment can involve multiple Healthcare Providers but it involves necessarily one single person.

One should bear in mind that a relator type is not as a UML association class \cite{guizzardi2008s}. The latter is identical to an instance of an association, i.e., an objectified n-tuple. A relator instead grounds those n-tuples. This is represented here by the dashed line connecting the \textit{treated by} material relation and the \textit{Treatment} relator type. This relation is called \textit{derivation} \cite{guizzardi2005ontological} and it connects tuples and relators in the following manner: a tuple $<x_1...x_n>$  is derived from a relator r if r \textit{mediates}\footnote{See the relation of \textit{mediation} defined at type level in Fig. \ref{fig:ontouml-model1}, which, as discussed in \cite{guizzardi2005ontological}, represents a type of multiple existential dependence relation.} (is existentially dependent on) $x_1...x_n$. Derivation, as any other association, can be subject to its own cardinality constraints. In this example, we can specify from how many instances of treatment can a tuple instantiating \textit{treated by} (e.g., $<John, TGH>$) be derived, i.e., how many times can a person be treated by the same Healthcare Provider.  

\subsection{Clarifying distinctions within types}
Here, again, bearing a treatment is not an essential property of people. In fact, treatments are established between healthcare providers and a subtype of unhealthy persons, which we call patients. A patient is an entity of kind person and, like Unhealthy Person, it is a dynamic type, i.e., no patient is necessarily a patient, and people can move in and out of the extension of patient without changing their identity. However, unlike Unhealthy Person, the dynamic classification condition here is not an intrinsic but a relational one, namely, a patient is a person that participates in at least one treatment by a healthcare provider. Dynamic and relational types specializing a kind are called \textit{Roles} \cite{guizzardi2005ontological,guarino2000formal}. Finally, this model makes explicit that Healthcare Provider here is considered a \textit{subkind} \cite{guizzardi2005ontological} of Organization, i.e., an organization whose distinguishing trait (and its essential nature) is that it treats patients. In other words, organizations that are Healthcare Providers are essentially such.

The original model of Fig.\ref{fig:simpleuml} occludes many fundamental aspects of this domain. In constrast, in the unpacked model of Fig.\ref{fig:ontouml-model1}, we can elaborate on the essential kinds of entities in these domains, their contingent phases, the roles they play, we can disambiguate cardinality constraints of multiple kinds, and we can explain why certain relations hold when they do. The latter model would also allow to go further specifying temporal properties of medical conditions, phases of the treatment relator (e.g., Active, Suspended, Finalized Treatment) as well as events governing the life-cycle of treatments (e.g., creation, termination, suspension, reactivation events), subkinds of treatment (e.g., Inpatient and Outpatient Treatment), roles of treatment (e.g., the role of Insured Treatment that a Treatment plays when covered by a Medical Insurance), etc. 

Now, and this is perhaps the most important take away message of this paper, the difference between the models in Figures \ref{fig:simpleuml} and \ref{fig:ontouml-model1} is not just one of expressivity but one of nature: the former has a \textit{merely descriptive nature}, focusing on describing truth-bearers; the latter has an \textit{explanatory nature}, focused on identifying and making explicit the underlying truthmakers. It is models of the latter sort that can properly support us in tasks of domain understanding, problem-solving, and meaning negotiation for tackling semantic interoperability challenges.    
\subsection{Considering alternative conceptualizations}
\begin{figure}
    \centering
    \includegraphics[width=.7\textwidth]{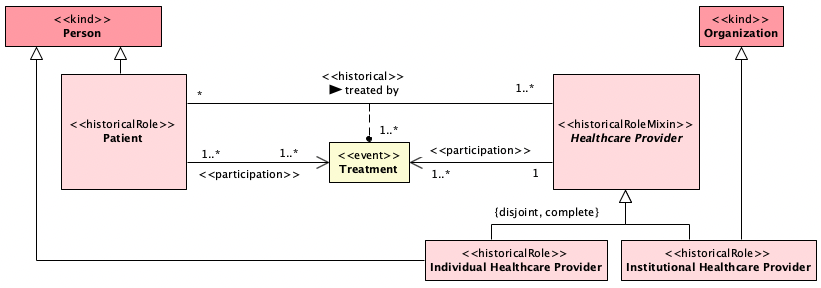}
    \caption{An alternative unpacking for the model of Fig.\ref{fig:simpleuml}.}
    \label{fig:ontouml-model2}
\end{figure}

To briefly illustrate how revealing the real-world semantics of a model can support semantic interoperability, let us rollback our unpacking exercise to the original model of Fig.\ref{fig:simpleuml}. In particular, let us focus on the \textit{treated by} relation. We can easily imagine a completely different conceptualization for (roughly) the same domain that would also be occluded by the original UML model of Fig.\ref{fig:simpleuml}. In an alternative conceptualization (represented by the model of Fig.\ref{fig:ontouml-model2}), the domain view is over treatments that have happened in a historical perspective. So, we have that: (i) treatments are not relators (i.e., bundles of commitments, claims, etc.) but they are events; (ii) since events can only exist in the past\footnote{Ontologically speaking, events are histories of changes in certain qualities of entities and, as such, they can only exist in the past. Alleged `future events' are either expectations (beliefs) or commitments about possible ocurrences of events of a given type. For an in depth discussion on these issues, one should refer to \cite{guizzardi2016ontological,guarino2022events}.}, the model acquires a historical semantics, i.e., being a Patient here is \textit{to have been treated by} a Healthcare Provider, or, in other words, it is to participate in a treatment event playing that role while the event was occurring; (iii) mutatis mutandis, the same can be said for being a Healthcare Provider\footnote{In OntoUML, ``roles" played by entities of a given kind while participating in an event are termed \textit{Historical Roles}; ``roles" that can be played by entities of multiple kinds are called \textit{Role Mixins}; naturally, ``roles" that can be played by entities of multiple kinds while participating in a event are termed \textit{Historical Role Mixins} \cite{guizzardi2005ontological,almeida2019events}.}; (iv) what makes true that a patient has been treated by a Healthcare Provider is that they both participate playing the appropriate ``roles" in at least one treatment event; (v) here, Healthcare Providers are not only Organizations but also individuals that provide care in treatment events.

By explaining what it means to be Patient, Healthcare Provider, Treatment, as well as the semantics of the \textit{treated by} relation, these alternative ontological unpackings of Fig.\ref{fig:simpleuml} can make explicit that the relations between the aforementioned types in the models of Figures \ref{fig:ontouml-model1} and \ref{fig:ontouml-model2} cannot be one of identity, simply because, despite being homonymous, these types belong to different ontological categories. Perhaps, the correct relation connecting Treatment in the latter model (i.e., treatment relators) and Treatment in the former (i.e. treatment events) is a relation of \textit{manifestation} \cite{guarino2015we,guizzardi2016ontological}, such that a treatment event manifests the commitments, claims, among other aspects, constituting that relator. As a consequence, since manifestation is a historical dependence relation, Patient and Institutional Healthcare Provider in the latter model would bear historical relations to Patient and Healthcare Provider in the former sense, e.g., someone playing a historical role of Patient in a treatment event is someone who played the role of Patient while mediated by the treatment relator from which that event derives.    
%is a relation of \textit{focus} \cite{guarino2015we}, in the sense that the relational aspects that constitute the relator are the main ``subject of change'' in the treatment event. Alternatively, we can have that a  treatment event is a manifestation of a treatment relator (i.e., the commitments, claims, etc. constituting that relator)%(i.e., a treatment relator while in an active phase)
%. In any case, a consequence of the latter interpretation is that, since manifestation is a dependence relation, hence, Patient and Institutional Healthcare Provider in the latter model would bear historical relations to Patient and Healthcare Provider in the former sense -- derived from this historical dependence between treatment in both senses. 

Fully analyzing these ontological relations between these two models is beyond what we intend to achieve here. An important point to highlight is 
% I would delete the words "to highlight"
that the original UML model collapses two very different conceptualizations involving some different ontological entities. Exposing these differences requires providing a proper explanation for the meaning of these entities.  

\subsection{Recognizing Patterns}

In the discussion above, we have employed the ontology-driven conceptual modeling language OntoUML in those ontological unpacking exercises. Although this is not strictly necessary, the task is much facilitated by using a modeling language that was designed to offer primitives that are based on true ontological distinctions, and methodological guidelines that force the modeler to search for and explicitly represent truthmakers. 

Additionally, as shown in \cite{ruy2017reference,zambon2017formal}, OntoUML is actually an \textit{ontology pattern grammar}, and its models are built and interpreted by employing \textit{ontology design patterns}.  These patterns, in turn, represent micro-theories of a background foundational ontology \cite{guizzardi2005ontological,ruy2017reference}. This is an aspect of this approach that is analogous to the idea of \textit{theoretical reduction} in the Unificatory approach\footnote{In the unificatory approach, explanations are built by employing certain general \textit{argument patterns} such that our knowledge of specific phenomena can be derived from the same set of basic patterns. In this sense, these basic patterns `unify our experience' \cite{weber2013scientific}. As Kitcher puts it: \textit{``Science advances our understanding of nature by showing us how to derive descriptions of many phenomena, using the same patterns of derivation again and again, and, in demonstrating this, it teaches us how to reduce the number of types of facts we have to accept as ultimate (or brute)''} \cite{kitcher1989explanatory}.  } to explanation \cite{kitcher1981explanatory}. 

For example, material relations are always modeled as instantiations of the \textit{Relator Pattern} in \cite{ruy2017reference}, which in turn, is a representation of a set of constraints composing a fragment of the theory of relations discussed in section \ref{analysis}. In that pattern, (in a nutshell) we have that if a material relation \textit{MR} is established between relata of types $T_1$ and $T_2$ then there is a Relator Type \textit{Rel} such that the latter bears a \textit{derivation} relation with \textit{MR}, and \textit{mediation} relations types $T_1$ and $T_2$. As a result, we have that, for each instance $<x_1,x_2>$ instantiating \textit{MR}, then: (i) $x_1$ instantiates $T_1$; (ii) $x_2$ instantiates $T_2$; (iii) there is a relator \textit{r} instantiating \textit{Rel} such that \textit{r} \textit{mediates} (and, hence, is \textit{existentially dependent} on) $x_1$ and $x_2$. Now, notice that this is a formal (i.e., domain-independent) theory, and that all domain material relations can be modeled by this pattern. Or, to rephrase it mimicking a unificatory jargon, domain phenomena are \textit{reduced to} (and, hence, explained in terms of) the background micro-theories represented by these patterns, such that instances of material relations such as \textit{married with}, \textit{enrolled in}, \textit{employed at}, \textit{president of} are \textbf{grounded} (or \textit{derived from}) instances of \textit{marriages}, \textit{employments}, \textit{enrollments}, \textit{presidential mandates}, respectively\footnote{Contrast this notion of ontological patterns with the notion of argument patterns in the unificatory approach \cite{weber2013scientific}. In a similar manner, these ontological patterns provide for a \textit{schematic structure} in which non-logical expressions (e.g., MR, $T_1$, $T_2$, Rel) can be replaced by domain-specific monadic types and relations following some \textit{filling instructions}.}.

\subsection{Visualizing possible interpretations}
Finally, there is another feature offered by the OntoUML ecosystem that we shall not fully explore here but that can have an important explanatory function, namely, its support for model validation via visual simulation. 
% I would start with "This feature allows ..." and then just put the citation at the end. 
As shown in \cite{benevides2010validating}, this feature allows one to visualize the possible \textit{interpretations} (in the logical sense) of an OntoUML model, that is, whether the conceptual model is ``saying on behalf" on the modeler what the modeler thinks it is saying, i.e., what the modeler intends it to say. As empirically demonstrated in \cite{sales2015ontological}, this is rarely the case, i.e., especially given the model semantics of the language, modelers can hardly imagine all the possible configurations satisfying the constraints of the model, for any model that is not trivial. 

\begin{figure}
    \centering
    \includegraphics[width=.5\textwidth]{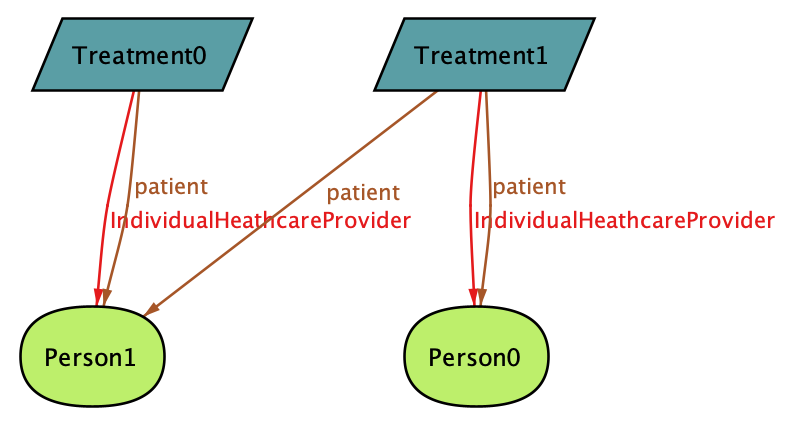}
    \caption{An unintended but possible interpretation for the model of Fig.\ref{fig:ontouml-model2}.}
    \label{fig:alloy}
\end{figure}

To offer one simple example, take the interpretation depicted in Fig. \ref{fig:alloy},  which is one of the possible automatically generated interpretations (i.e., logical model) for the model of Fig.\ref{fig:ontouml-model2}. This instance model makes clear that the same individual ($Person_1$) can be a patient in a treatment ($Treatment_1$) and healthcare provider in another treatment ($Treatment_0$). However, in the absence of additional constraints, 
% it is also clear that ... 
it also makes clear 
that the same person (again, $Person_1$) can be patient and healthcare provider in the very same treatment ($Treatment_0$), which might not be an interpretation intended and previously identified as a possibility by the creator of a model such as the one in Fig.\ref{fig:ontouml-model2}\footnote{In fact, as shown by \cite{sales2015ontological}, this is a common modeling anti-pattern. The visualization of these possible instances is offered there as an explanation for a model such as this one, containing an instance of that anti-pattern.}. This case can happen because treatments there involve patients and healthcare providers, both of which can be instantiated by instances of the kind Person (in case of an Individual Healthcare Provider for the latter) and, hence, they can be instantiated by the very same person. In summary, by providing an automatic way to visualize the actual ontological commitment of a description, this functionally provides a key support for model explanation. 

\section{Explanations in Philosophy, Science, and XAI}
 \label{more}
\subsection{Philosophical and Scientific Explanation}

There is a vast literature in philosophy of science proposing a myriad of types of explanation. These include Nomological-Deductive \cite{hempel1965aspects}, Teleological \cite{cohen1950teleological}, Pragmatic \cite{hempel1965aspects}, and Unificatory \cite{kitcher1981explanatory} explanations, among others. There is also a rich literature in philosophy on the topic \cite{brenner2021metaphysical,smith2007truthmaker}. Reviewing these bodies of literature would go far beyond the goals of this paper. In the sequel, we shall briefly compare some aspects of the approach discussed here with some of the classical approaches in philosophy. 

Ontological unpacking is directly relatable to the notions of grounding and truthmaking in the philosophical literature on metaphysical explanations. However, it also bears some interesting relations to other forms of explanation. Moreover, as previously mentioned, it is akin in spirit to the notion of \textit{ontological explanation} defended by T. Y. Cao \cite{cao2004ontology}, as both approaches defend that the source of explanatory power of an explanans of a syntactical artifact (here represented by another syntactical artifact that is the unpacked domain model) comes from identifying ontological entities (e.g., objects, events, dispositions, qualities, relators) and their ties. Moreover, ontological unpacking (as conducted here) makes use of a number of  design patterns \cite{ruy2017reference} representing formal (i.e., domain-independent) micro-theories constraining the relations between these ontological entities. As a result, in the explanation process, domain phenomena are interpreted in terms of these patterns. This mechanism resembles the mechanism of explanation as theoretical reduction proposed by a unificatory account. Furthermore, as discussed in \cite{romanenko2022towards}, the notion of explanation advocated here is also very much aligned with a \textit{pragmatic approach to explanation}\footnote{In this approach proposed by Van Fraasen \cite{van1977pragmatics}, an explanation is an answer to a \textit{request for explanation}, the latter being characterized by a \textit{topic}, as well as a \textit{contrast-class} and a \textit{relevance} relation, so that a topic is true (as opposed to the alternatives in the contrast-class being true) because of a given reason R that is consider relevant in a given context to explain the topic. Both contrast-class and relevance relation make the \textit{explanatory relevance} of explanations (i.e., of reasons) dependent on the context of an explanation-seeker. The term `pragmatic' refers to seeing explanations as tools in a toolbox, as opposed to describing some essential properties of entities in reality \cite{weber2013scientific}. } \cite{van1977pragmatics} in at least two senses, which are briefly summarized in the sequel.

Firstly, truthmakers should support answers to \textit{requests for explanation} \cite{van1977pragmatics}%\footnote{In \cite{van1977pragmatics}, Van Fraasen defends that a theory of explanation is a theory of why-questions.}
. In the ontology engineering literature in Computer Science, the models representing domain ontologies are typically designed to answer so-called Competency Questions (CQs) \cite{gruninger1995role}. These, however, are typically formulated as mere \textit{lookup queries} that navigate the structure of the model \cite{bezerra2013evaluating}. In contrast, we defend 
% I would say "I propose.." 
that if CQs are to be of any use in guiding the construction of these models, they should be formulated as genuine context-dependent requests for explanation about the domain\footnote{Even contrastive questions in Van Fraasen's sense can play a role here. If we refer to the model of Fig.\ref{fig:ontouml-model1}, the question ``Why is a person treated by a given healthcare provider?" can be interpreted both as: (a) ``Why is a person treated by a given healthcare provider as opposed to another healthcare provider?''; or (b) ``Why is a person treated by a given healthcare provider as opposed to not being treated?''. In the case of (a), the relevant explanation would be articulared in terms of the Treatment relator, e.g., ``John is treated by TGH because there is a treatment relator binding them''; in (b), in contrast, the relevant explanation would be articulated in terms of Pathological Health Condition modes, e.g., ``John is treated by TGH because he has Diabetes''.}. %This is a topic that we will systematically investigate in a future paper. 

Secondly, in a pragmatic view, explanations should always be tuned to the relevant characteristics of the explanation-seeker (e.g., background, objectives, etc.). In that sense, to explain is to ``clarify, to make it flat/plain"\footnote{This is emphasized, for example, in the Dutch terms for explanation \textit{verklaring} (to clarify) and \textit{uitleg} (to lay it out).}, i.e., to reduce complexity. This aspect of explanation goes exactly in the inverse direction of unpacking. One one hand, the process of ontological unpacking reveals the distinctions that make up the conceptualization underlying domain models, thus, typically yielding larger and more detailed models. On the other hand, as  discussed in \cite{romanenko2022towards}, this activity must be complemented by complexity management tools for, e.g., model \textit{abstraction} (or summarization), \textit{modularization}, and \textit{viewpoint extraction} \cite{figueiredo2018breaking,guizzardi2021automated,romanenko2022abstracting,guizzardi2019ontology}, which could be adapted to reduce the complexity of the model with an explanation-seeker or a set of requests for explanation in mind. This is another direction of work 
%that intend to pursue with colleagues as
that is an extension of these methods.  

\subsection{Explainable Artificial Intelligence (XAI)}

Recently, there has been a growing interest in Computer Science, in particular, in Artificial Intelligence, on the topic of explanation \cite{dwivedi2023explainable}. It is motivated by the growing popularity of sub-symbolic AI, which is  increasingly being applied  to situations where it could conflict with human values, goals and norms, and the intrinsic inscrutability of some of its techniques. Explanation even features at the European Union General Data Protection Regulation (GDPR) in terms of an algorithmic decision-making accountability demand, the so-called \textit{Right to Explanation} \cite{kaminski2021right}. For this reason, as computer scientists, we cannot shun discussing the  topic of so-called Explainable Artificial Intelligence (XAI). A full discussion on the topic would of course require a separate paper, so we shall be brief, just touching the two main approaches to XAI, namely the \textit{Interpretability Framework} and the \textit{Explainability Framework}. 

In the so-called Interpretability Framework, or Complete Model View, to explain is interpreted as ``to produce a symbolic artifact'' (e.g., a decision-tree, perhaps extended by a ``lightweight ontology''\footnote{The term ``ontology" is used in this context as a logical specification, typically encoded in a sort of computational logics termed OWL (Web Ontology Language). As Guizzardi has argued elsewhere \cite{guizzardi2007conceptualizations}, the acronym is a misnomer given that OWL is just a logical language and, hence, it is ontologically neutral. Likewise, the logical specifications represented in OWL are very much focused on supporting automated reasoning while retaining certain computational properties, and are seldom genuine ontological representations in the sense we have used in section \ref{semantics-ontology}.}\cite{confalonieri2019ontology}) 
that is supposed to spell out completely the decision-making process carried out by the AI component of the system (currently, typically a black-box). Now, suppose we are able to produce a large rule-set that is formally equivalent to the information processes carried out by a Neural Network (NN)%\footnote{It is unsurprising that this can be done, since a NN is ultimately a very complex function, and finite functions can be written as sets of if-then rules%(i.e., \textit{f(a) = b} is equivalent to \textit{If a then b)}. See \cite{aizawa1994representations} for a broader discussion.}
. In what sense is this large rule-set an explanation for those processes? Letting aside the issue of whether the original decision-making processes of the network can be reconstructed in a cognitively tractable manner by a human agent looking at a large set of if-then rules, there is a mistaken assumption that is even more fundamental here: reducing the activity of explaining to an activity of generating a symbolic model presupposes that symbolic models are self-explanatory! In fact, these symbolic models are often called by the XAI community \textit{``inherently interpretable models"}! \cite{sudjianto2021designing}. 

However, as illustrated in Section \ref{analysis}, there is nothing `inherently interpretable' even in symbolic models with a clear formal semantics describing small, simple and familiar domains as the one illustrated in Fig.\ref{fig:ontouml-model1}. There is still a challenging road from symbolic artifacts (such as rule-sets/decision-trees and computational ontologies extending them) to explanations that really reveal the ontological commitments of information structures and processes and that properly answer meaningful \textit{requests for explanation}, 
offering some systematic support for that. Additionally, in our view, \textit{explanation as accountability} requires that we properly identify the truthmakers of propositions about which we should explicitly collect evidence \cite{romanenko2022towards}, and that this evidence has a clear, precise, unambiguous semantics. 

Now, let us also very briefly comment on the \textit{Explainability Framework} (also called \textit{Partial-Model Approach}). This approach aims at emulating the decision-making process carried out by a black-box (nowadays, typically, a Deep Neural Network) without opening the black-box. In this camp, a technique that gained prominence for allegedly preserving some social and psychological aspects of explanation \cite{miller2019explanation} is the one based on so-called \textit{counterfactual explanations} \cite{wachter2017counterfactual}. According to this technique, to explain is to produce a minimal set of changes to the input data that would yield the desired decision. To cite an example from \cite{wachter2017counterfactual}: \textit{(CE) “You were denied a loan because your annual income was £30,000. If your income had been £45,000, you would have been offered a loan.”} In this example, the decision of why someone was denied a loan is explained in terms of a counterfactual situation in which that loan would have been granted. The characterizing features differentiating the two situations are in a sense the explanans. 

However, as demonstrated by Browne and Swift \cite{browne2020semantics}, generally speaking, the characterization of explanations in this way makes them nearly indistinguishable from \textit{adversarial examples}. As they put it, generating ``counterfactual explanations consisting only of semantically dense and contextually relevant dimensions in the network’s feature space" is what it would take to differentiate the two cases (i.e., account for what they call the explanatory divide). However, in order to do that, we would need to be able to \textit{reveal the semantics} of hidden network units (`hidden neurons') \cite{browne2020semantics}. In fact, their phrase \textit{``there can be no explanation without semantics"} makes for a motto to which we could not agree more. 

Now here is one problem: as admitted by Browne and Swift, we do not know how to do this, i.e., to reveal the semantics of hidden units. Here is a much bigger problem: as discussed by Saba \cite{saba}, \textit{``representations in NNs are not really `signs' that correspond to anything interpretable — but are distributed, correlative and continuous numeric values that on their own mean nothing that can be conceptually explained. Stated in yet simpler terms, the subsymbolic representation in a NN does not on its own refer to anything conceptually understandable by humans (a hidden unit cannot on its own represent any object that is metaphysically meaningful"})\footnote{Saba \cite{saba} goes on making the case ``that is precisely why explainability [taken as inference in reverse] in NNs cannot be achieved, namely because the composition of several hidden features is undecidable—once the composition is done (through some linear combination function), the individual units are lost". }. In other words, the `semantics' referred to by Browne and Smith is not and cannot be the type of real-world semantics we have been talking about here%\footnote{Making justice to ontological commitments in the sense we have been discussing here requires the ability to deal with intensional notions (see \cite{guarino1998formal} but also \cite{guizzardi2015logical}). In a complementary direction, Saba (also relying on Fodor \& Physylyn \cite{fodor1988connectionism}) makes the point that extensional models such as NN cannot model essential intensional aspects %such as productivity, compositionality and systematicity, which are essential characteristics of not only conceptual representations but also 
%of mental processes.}
. However, even if this would be at all possible for the general case, i.e., building counterfactual explanations in this sense in terms of domain dimensions (e.g., annual income) there is still the very hard task of interpreting the semantics of these dimensions, especially in social and legal domains. As it is very often the case, legal disputes boil down to exactly determining the very subtle semantics of domain terms\footnote{One excellent example mentioned by Pinker in \cite{pinker2007stuff} is the case of the terrorist attacks on the WTC on Sept 11, 2001. Deciding whether the tenants of the towers would be eligible to a 3.5 or 7 Billion US\$ compensation relied on the semantics of the term `occurrence' and, ultimately, on ontological commitments towards identification and unity conditions for events.}, and we do know how this could be done without serious ontological analysis, conceptual clarification and meaning negotiation work. 

\section{Final Considerations}\label{final-considerations}

We have briefly discussed some of the relations between different senses of semantics, of ontology, and of explanation. In particular, we focused on a notion of semantics termed \textit{real-world semantics} (also dubbed \textit{ontological semantics}), which is about the relation between a description (in particular, a conceptual model) and an \textit{ontology}, i.e., a theory about the kinds of categories and their ties that are assumed to exist if that description is true.  

All representations that have real-world semantics (i.e., which are not mere formal structures) make an \textit{ontological commitment}. That is, they commit to a particular ontology of the domain. The activity of precisely identifying and analyzing the entities constituting that ontology is termed \textit{ontological analysis}. Doing this requires the support of \textit{Ontology} -- as a discipline providing a body of theories, methods and conceptual tools, and we must do this in a systematic and precise manner. First, because the way from description to revealed ontology is a tortuous one, and full of traps \cite{varzi2007language}. Second, because of challenges and criticality that semantic interoperability in different manifestations poses to us.

For example, in Big Data scenarios, the management cost and effort is concentrated on dealing with the \textit{variety} (including semantic variety) of data sources and their \textit{veracity} \cite{bean2016variety,jaulent2018semantic}. Moreover, the quality of data analysis results is ultimately dependent on how well that data maybe considered as ``a theory of the real-world" (to quote Mealy). As stated by Hannigan \cite{hannigan2015close}: ``It is not heaps of transactional data that make an inquiry scientific. Being scientific is an effect of work done to establish stable, quantifiable concepts ... the \textit{concepts are a prerequisite for the existence of the data”} (our emphasis). Or, as expressed by Ned Block \cite{block2014consciousness}: ``high resolution data are of no use without a theory. When we have substantive theories— together with the sophisticated concepts... testing these theories may require Big Science. But we cannot expect the theories and concepts to somehow emerge from Big Science". In summary, we need to take semantics and ontology very seriously if we are going to offer trustworthy support for properly producing, interpreting and semantically interoperating a multitude of data silos in critical domains. 

Revealing the ontology underlying a given domain model is a type of \textit{explanation} that we have termed \textit{ontological unpacking}. The idea of ontological unpacking is to identify and make explicit the \textit{truthmakers} of the propositions represented in that domain model, i.e., the entities in the world that make those propositions true, and from which logico-linguistic constructions (such as derived types, attributes, associations) are derived. 

We hope to have succeeded in making the case that the difference between classical conceptual models and ontology-driven conceptual models is not just one of expressivity, but instead one of nature. While the former typically focus on a merely descriptive and information structuring function for data processing applications, thus, typically concentrating on the representation of truth-bearers, the latter focus on an \textit{explanatory function} aiming at revealing and explicitly characterizing the relevant truthmakers out there in the world. As put forth by Van Fraasen's \cite{van1977pragmatics}: \textit{``Traditionally, theories are said to bear two sorts of relation to the observable phenomena: \textbf{description} and \textbf{explanation}. Description can be more or less accurate, more or less informative; as a minimum, the facts must `be allowed by the theory' (fit some of its models), as a maximum the theory actually implies the facts in question. But in addition to a (more or less informative) description, the theory may provide an explanation. This is something `over and above' description; for example, Boyle's law describes the relationship between the pressure, temperature, and volume of a contained gas, but does not explain it - kinetic theory explains it...even if two theories are strictly empirically equivalent they may differ in that one can be used to answer a given request for explanation while the other cannot.''} That said, we had no ambition here to contribute to the vast philosophical literature on scientific and metaphysical explanation. 
The task was a much more modest one, but one that a truly scientific approach to data cannot shy away from.    

As a final note, Babic and colleagues \cite{babic2021beware} propose a different view on the problem of explanation, namely, one should replace trying to explain artifacts by building trust in the process that produces these artifacts. For example, we accept prescriptions from our doctors, and travel in planes flown by our pilots without asking these individuals to explain to us how they do what they do. Sure, but we can collectively trust the process that produce medicines, airplanes, physicians and pilots because there are individuals in these processes that do have access to semantically transparent and precise explanations of concepts (and theories based on them) that make these processes possible. In our view, we can only build trustworthy systems if they are based on trustworthy explicit representations, and the latter can only be achieved with the proper tools for dealing with semantics and ontology. To put it simply: ``there is no explanation without semantics" \cite{browne2020semantics,saba}, there is no semantics without ontology, and, there should be ``no ontology without Ontology" \cite{varzi2019carnapian}. 
\\\\
\textbf{Acknowledgements.} We are grateful to Roberta Ferrario, Riccardo Baratella, Mattia Fumagalli, Veda Story, Renata Guizzardi, Alberto Garcia Simon, Anna Bernasconi, Oscar Pastor, Erik Proper, Barend Mons, Walid Saba, Diego Calvanese, and Elena Romanenko, for many insightful discussions and comments that helped to improve the content this paper. We are also greateful to Tom Gauld and Michiel Straathof for kindly granting their permissions to use their artwork in Figures 1 and 2. 

\bibliography{sn-bibliography}

\end{document}